\documentclass{article}

\usepackage[nonatbib,preprint]{neurips_2025}

\usepackage[T1]{fontenc}    
\usepackage[colorlinks, 
]{hyperref}
\usepackage{url}            
\usepackage{booktabs}       
\usepackage{amsfonts}       
\usepackage{nicefrac}       
\usepackage{microtype}      
\usepackage{xcolor}         

\usepackage{multirow}
\usepackage{tabularx}
\newcolumntype{C}{>{\centering\arraybackslash}X} 
\usepackage{graphicx}
\usepackage[square,numbers]{natbib}
\usepackage{amsmath}
\usepackage{xspace}
\newcommand{\method}{Lumen\xspace} 

\title{
  \parbox{1cm}{\centering\includegraphics[height=1.4em]{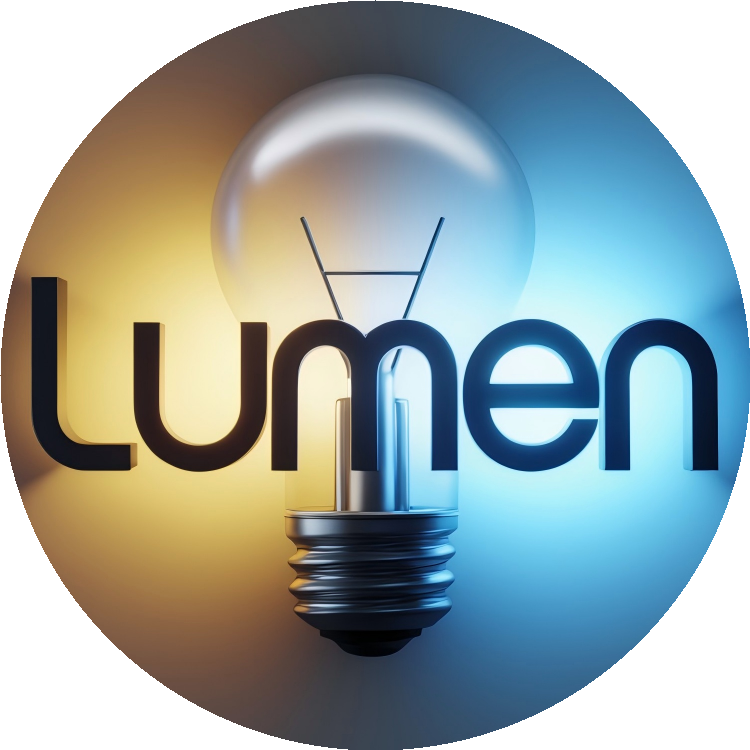}}
  \method: Consistent Video Relighting and \\ Harmonious Background Replacement \\ with Video Generative Models
}

\author{%
Jianshu Zeng$^{1,3*}$, Yuxuan Liu$^{2*}$, Yutong Feng$^{2\dagger }$, Chenxuan Miao$^4$, Zixiang Gao$^1$, \\
\vspace{+3pt}
\textbf{Jiwang Qu$^5$, Jianzhang Zhang$^5$, Bin Wang$^2$, Kun Yuan$^{1\ddagger}$} \\
$^1$Peking University, $^2$Kunbyte AI, $^3$University of Chinese Academy of Sciences, \\
\vspace{+3pt}
$^4$Zhejiang University, $^5$Hangzhou Normal University \\
\texttt{\{zengjianshu.ai,liuyuxuanuestc,fengyutong.fyt,weiyuchoumou526\}@gmail.com,} \\
\texttt{2401210062@stu.pku.edu.cn, 2024112013037@stu.hznu.edu.cn, }\\
\vspace{+3pt}
\texttt{zjzhang@hznu.edu.cn, binwang393@gmail.com, kunyuan@pku.edu.cn} \\
$^* $Equal Contribution. $^\dagger $Project Leader. $^\ddagger $Corresponding Author.
}

\begin{document}

\maketitle

\begin{figure}[!h]
  \centering
  \includegraphics[width=1.0\linewidth]{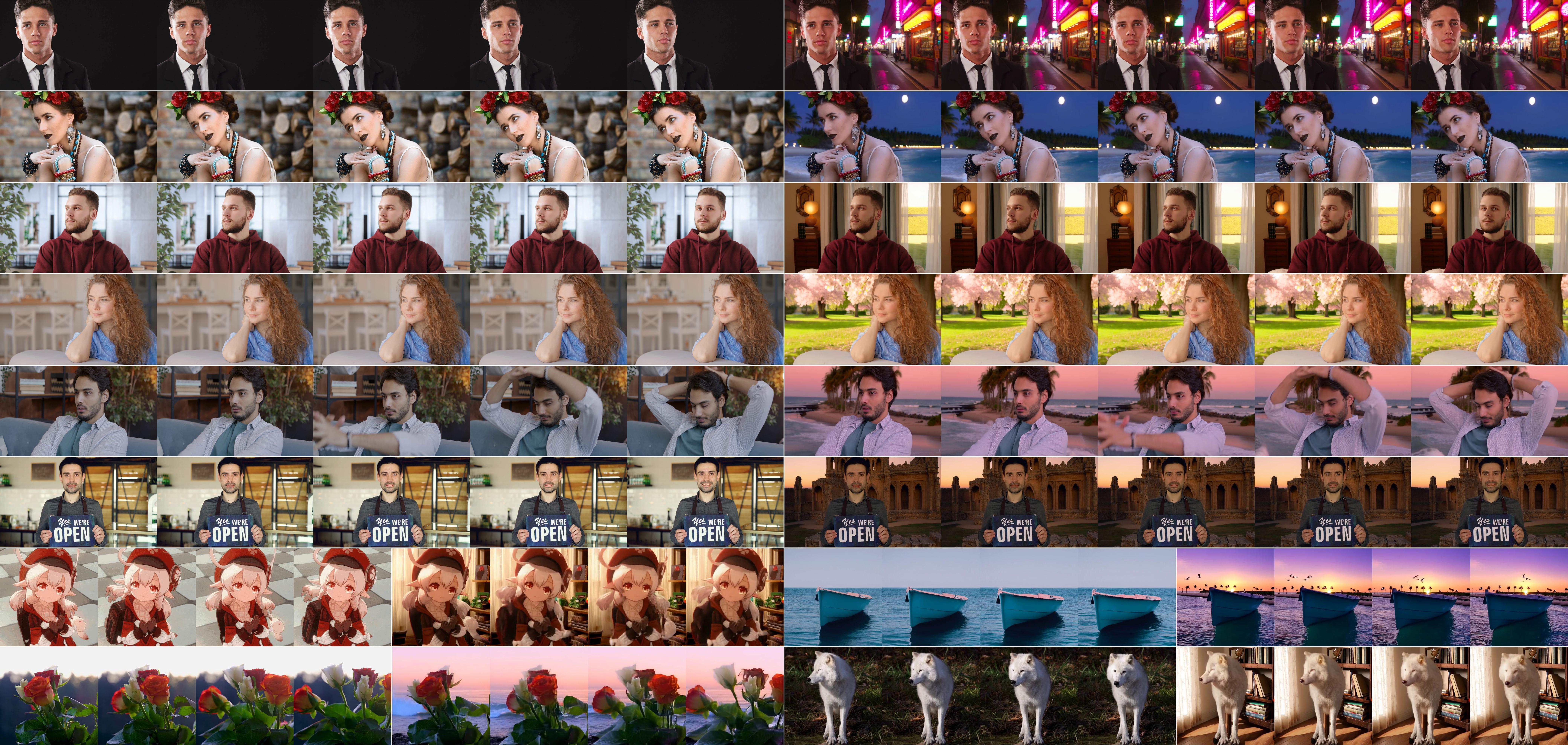}
  \caption{ \textbf{Video relighting results of \method}, where the left sequence shows sampled key-frames of input video and the right sequence shows relighted video. \method achieves harmonious video relighting with consistent foreground preservation for various characters, scenarios and domains. Our project page: \url{https://lumen-relight.github.io/}}
  \label{fig:teaser}
\end{figure}

\begin{abstract}
Video relighting is a challenging yet valuable task, aiming to replace the background in videos while correspondingly adjusting the lighting in the foreground with harmonious blending.
During translation, it is essential to preserve the original properties of the foreground, \textit{e.g.,} albedo, and propagate consistent relighting among temporal frames.
While previous research mainly relies on 3D simulation, recent works leverage the generalization ability of diffusion generative models to achieve a learnable relighting of images.
In this paper, we propose \textbf{\method}, an end-to-end video relighting framework developed on large-scale video generative models, receiving flexible textual description for instructing the control of lighting and background.
Considering the scarcity of high-qualified paired videos with the same foreground in various lighting conditions, we construct a large-scale dataset with a mixture of realistic and synthetic videos.
For the synthetic domain, benefiting from the abundant 3D assets in the community, we leverage advanced 3D rendering engine to curate video pairs in diverse environments.
For the realistic domain, we adapt a HDR-based lighting simulation to complement the lack of paired in-the-wild videos.
Powered by the aforementioned dataset, we design a joint training curriculum to effectively unleash the strengths of each domain, \textit{i.e.,} the physical consistency in synthetic videos, and the generalized domain distribution in realistic videos.
To implement this, we inject a domain-aware adapter into the model to decouple the learning of relighting and domain appearance distribution.
We construct a comprehensive benchmark to evaluate \method together with existing methods, from the perspectives of foreground preservation and video consistency assessment.
Experimental results demonstrate that \method effectively edit the input into cinematic relighted videos with consistent lighting and strict foreground preservation.

\end{abstract}

\section{Introduction}

Video relighting with harmonious background replacement has attracted tremendous efforts~\cite{li2014free,guo2019relightables,zhang2021neural,lin2024edgerelight360,zhang2024lumisculpt,RelightVid_fang2025,light_a_video_zhou2025}, due to its widespread application in various scenarios, \textit{e.g.,} filmmaking and e-commerce.
Given the input video with foreground segmentation mask and the input condition, the objectives of video relighting can be formulated as two-fold: \textbf{1.} relighting the foreground while preserving the foreground with its intrinsic attributes, \textit{e.g.,} albedo and texture. \textbf{2.} replacing the background while harmonizing the lighting of foreground and background based on the input condition, \textit{e.g.,} text description or background image.
Traditional methods generally address this task with 3D-based physical simulations, such as spherical harmonics and light transport simulation.
More advanced studies introduce learning-based frameworks for editing the illumination of images~\cite{sun2019single,zhou2019deep,pandey2021total,yeh2022learning,kim2024switchlight} and videos~\cite{zhang2021neural,lin2024edgerelight360,choi2024personalized}, controlled by HDR map or Spherical Harmonic (SH) lighting.
Inspired by the recent success of large-scale generative models~\cite{stablediffusion,sdxl,flux2024}, it has been explored to unify the relighting task with end-to-end diffusion-based generation~\cite{iclight,light_a_video_zhou2025,RelightVid_fang2025}.
These methods leverage the generalization capabilities of diffusion models to achieve flexible lighting control with textual descriptions or background images, and can be widely-applied on ``in-the-wild'' data for broader real-world applications.

Despite the aforementioned advancements, existing works mainly concentrate on the image domain.
When confronting with the video input, it would be more challenging to propagate the relighting modification across all frames with temporal consistency. 
Extension to the video relighting is restricted by the scarcity of a large-scale dataset containing high-qualified and foreground-aligned pairs of videos.
We summarize all the necessary constitution of such a dataset as: \textbf{1.} video pairs with physically-aligned foreground and distinguished background conditions, \textbf{2.} corresponding textual descriptions of each video, especially on the background and lighting conditions, \textbf{3.} binary mask videos that segment the foreground.

In this paper, we construct a large-scale video relighting dataset to complement the scarcity, consisting of paired video samples from two domains:
For the \textit{synthetic} domain, powered by 3D rendering framework like the Unreal Engine~\cite{ue5}, we could obtain videos with fixed foreground subject and animation in various environments, which strictly follow the physical rules to preserve the foreground. However, models directly optimized on the 3D pairs suffer from the domain-gap problem, and fail to generalize to real videos.
For the \textit{realistic} domain, inspired by the random degradation strategy in IC-Light~\cite{iclight}, we prepare artificial paired videos by extracting the video normals~\cite{normalcrafter_bin2025} and simulating various lighting with random HDR maps. The weakness of the realistic pairs is that the foreground-preservation is not guaranteed, and there lack dynamics of light and shadow effects between videos.

Equipped with the above dataset, we presents \textbf{\method}, a video relighting model developed on large-scale video generative models~\cite{kong2025hunyuanvideo,wan2025,VideoX_fun2025} with hyper-realistic aesthetics and consistent temporal motions for video generation.
To effectively leverage the constructed dataset and benefit the strengths from each domain, we propose a multi-domain joint training curriculum in \method. 
In detail, we inject a style adapter into the model, serving to generate 3D-style videos with empty control signal.
Then the model is trained with mixed sampling of the realistic and synthetic data, where we activate the adapter only for the forward process of synthetic data.
During the inference, the adapter is deactivated to avoid introducing artificial quality into the edited video.
Throughout this way, \method decouples the style distribution of multiple domains and achieves natural output. 

We present a comprehensive benchmark for evaluating the video relighting performance, consisting of both realistic and synthetic videos as input.
Besides existing metrics regarding the similarity measurement and quality assessment~\cite{huang2024vbench}, we propose a new evaluation method, termed \textit{intrinsic consistency}, to measure the similarity after transferring both videos into uniform illumination.
Extensive experiments demonstrate the effectiveness of \method compared with existing methods, showing consistent video relighting and foreground preservation among video frames.

\section{Related Work}

\subsection{Video Generative Models}
The success of diffusion model~\cite{ddpm} on image generation has inspired the extension to video generation~\cite{video_diffusion_models}.
Pioneer works like Make-A-Video~\cite{make_a_video} design cascaded architecture to handle larger resolution generation.
Inspired by the latent diffusion model~\cite{stable_video_diffusion} on images, extensive researches achieve end-to-end video generation with convolutional networks for text-to-video~\cite{wang2023modelscope,emu_video,bar2024lumiere} and image-to-video~\cite{guo2023animatediff,chen2024livephoto} generation.
More recently, Sora~\cite{sora} demonstrates the potential of scaling diffusion transformers (DiT)~\cite{dit} to achieve significant advancement on the aesthetics, dynamics and smoothness of generated videos.
Following Sora, tremendous efforts explore the variants of DiT with various text encoders to achieve better generation quality and prompt following~\cite{yang2025cogvideoxtexttovideodiffusionmodels,kong2025hunyuanvideo,wan2025,bao2024vidu}.
The advancement of fundamental video generative models lays foundation to diverse downstream tasks, \textit{e.g.,} controllable video generation and video editing. 

\subsection{Image Relighting}

Image relighting aims to modify the lighting of objects in a static image under different illumination conditions. 
Early approaches primarily rely on physical illumination models~\cite{barron2014shape} or deep learning-based methods with specific lighting representations~\cite{sun2019single,zhou2019deep}.
Recently, the success of diffusion models has led to significant advancements in image relighting quality.~\cite{ddpm,cha2025text2relight,neuralGaffer_jin2024,DiFaReliPlus_2025,iclight}.
Some works mainly focus on the portrait relighting~\cite{DifFRelight,DiFaReliPlus_2025} or object relighting~\cite{relightingFromASingleImage,neuralGaffer_jin2024}, and they are limited to specific scenarios and lack the ability to control the lighting and background through text. 
Beyond that, some works explore the use of large-scale synthetic data to train diffusion models for text-guided and image-guided image relighting.
Text2Relight~\cite{cha2025text2relight} leverages a generative diffusion model with auxiliary task augmentation to enable text-guided portrait relighting from large-scale synthetic data.
IC-Light~\cite{iclight} introduces a light transport consistency loss and trains on a comprehensive dataset of rendered and augmented real-world images.
These advancements in image relighting provide essential foundations for extending the task to the temporal domain.

\subsection{Video Relighting}

Video relighting aims to modify the lighting of the dynamic foreground object in a video under different background conditions.
Current methods on portrait video relighting~\cite{at_home_light_choi2024,highfidelity_guo2025,lux_post_facto_mei2025,comprehensive_relighting_wang2025} often rely on the lighting source or the HDR image as the input, which limits the flexible application of the model.
Besides, the background in the generated videos is usually fixed, which leads to unnaturalness of the whole video.
More recently, inspired by image relighting advancements, some works have attempted to adapt image relighting models to video tasks for more flexible control and better results.
Light-A-Video~\cite{light_a_video_zhou2025} adapts the IC-Light by introducing Consistent Light Attention and Progressive Light Fusion strategies, while its train-free approach limits overall quality.
RelightVid~\cite{RelightVid_fang2025} inflates a pre-trained 2D image relighting model into a 3D U-Net with temporal attention layers, but struggles with foreground preservation and lighting consistency due to the limited quality of in-the-wild data and the inherent limitations of image relighting models.
Video relighting remains challenging due to the need for high-quality paired videos with different lighting conditions, and models that can effectively learn the lighting harmonization while maintaining the foreground and background consistency.

\begin{figure}[htbp]
  \centering
  \includegraphics[width=1.0\linewidth]{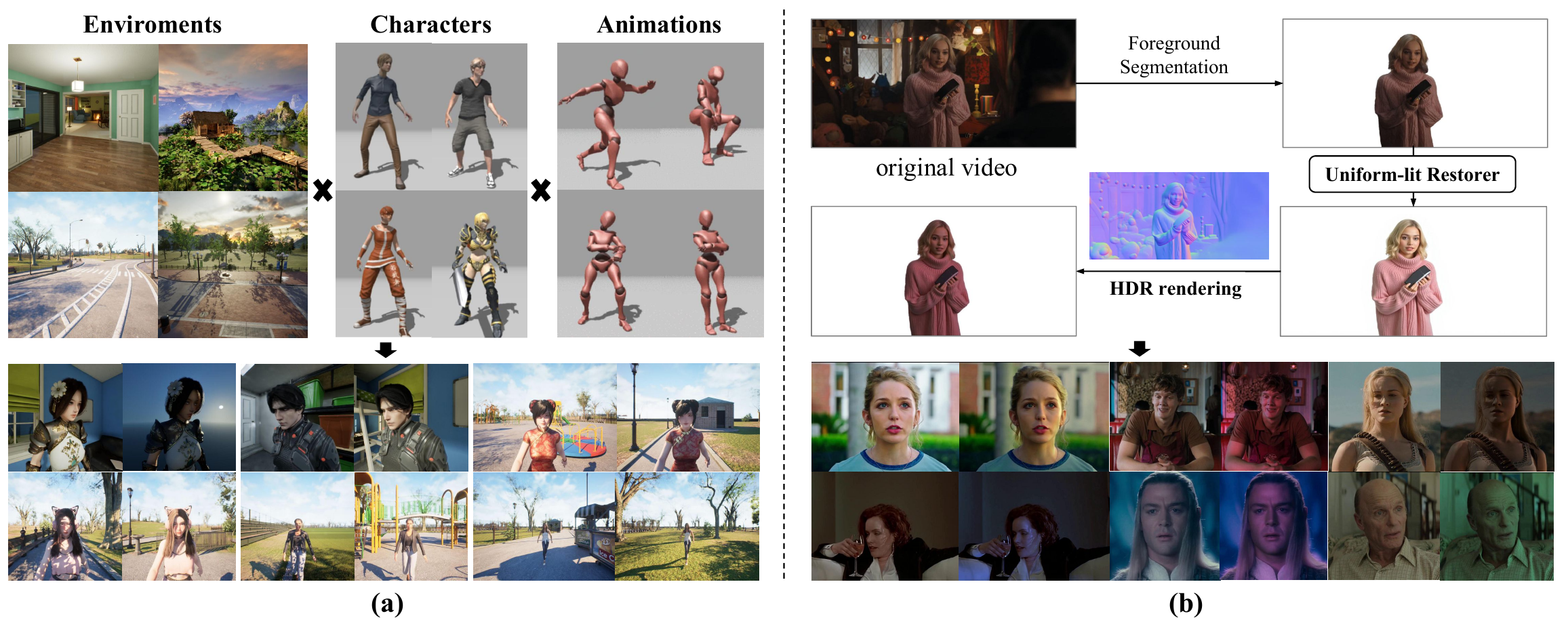}
  \caption{\textbf{The data preparation and examples of two domains.} (a) The 3D rendered data combines various environments, characters and animations to form paired videos with aligned foreground. (b) The realistic videos are transformed into uniform-lit appearance and rendered with HDR-based relighting.}
  \label{fig:dataset prepartion}
\end{figure}

\section{Method}

\subsection{Dataset Construction} 

In this section, we present the preparation pipeline for the paired video relighting dataset, which can be divided into the 3D rendered synthetic data, and the realistic in-the-wild data.
The dataset targets to serve a diverse range of lighting conditions and backgrounds to facilitate the training of video relighting.
\ref{fig:dataset prepartion} illustrates the pipeline and example pairs of our dataset.

\paragraph{\textbf{3D Rendered Data.}}
\label{sec:3D Rendered Data}
The rapid development of computer graphics enables us to render cinematic videos with precise control on the animation and environment.
Specifically, we leverage the advanced 3D rendering engine, \textit{i.e.,} the Unreal Engine 5 (UE5)~\cite{ue5}, to render video relighting pairs based on diverse 3D assets in the community. The data preparation can be summarized as follows:

\textit{3D Assets Collection:}
We first collect the necessary assets for rendering, including environments, characters (human or non-human objects), and animations from open-sourced websites, \textit{e.g.,} Fab~\cite{fab}, MetaHuman~\cite{metahuman}, Mixamo~\cite{mixamo} and other publicly available resources. 
The environments contain various lifelike indoor and outdoor scenarios with diverse lighting and background conditions, which are then split into small scenes for rendering.
The characters are generally equipped with sets of animations to simulate different subject motions.

\textit{Random Shot Generation:}
To obtain a valid shot focusing on the foreground, we first randomly sample a character and assign an animation towards it.
The animations can be divided into in-place animation and movement across the scene (\textit{e.g.,} walking).
The camera is set to be permanently oriented towards the character, while also moving in pre-defined camera trajectories, such as zoom-in and zoom out.
By combing different characters, animations and camera movements, we could obtain a large scale of valid shots in diverse distribution.
Another notable strength of rendered data is that we can automatically obtain the foreground mask video by setting specific texture to the character in the rendering engine.
It avoids to use segmentation parsers to detect the foreground, which could generate unstable masks for varying reasons.

\begin{figure}[htbp]
\centering
\includegraphics[width=1.0\linewidth]{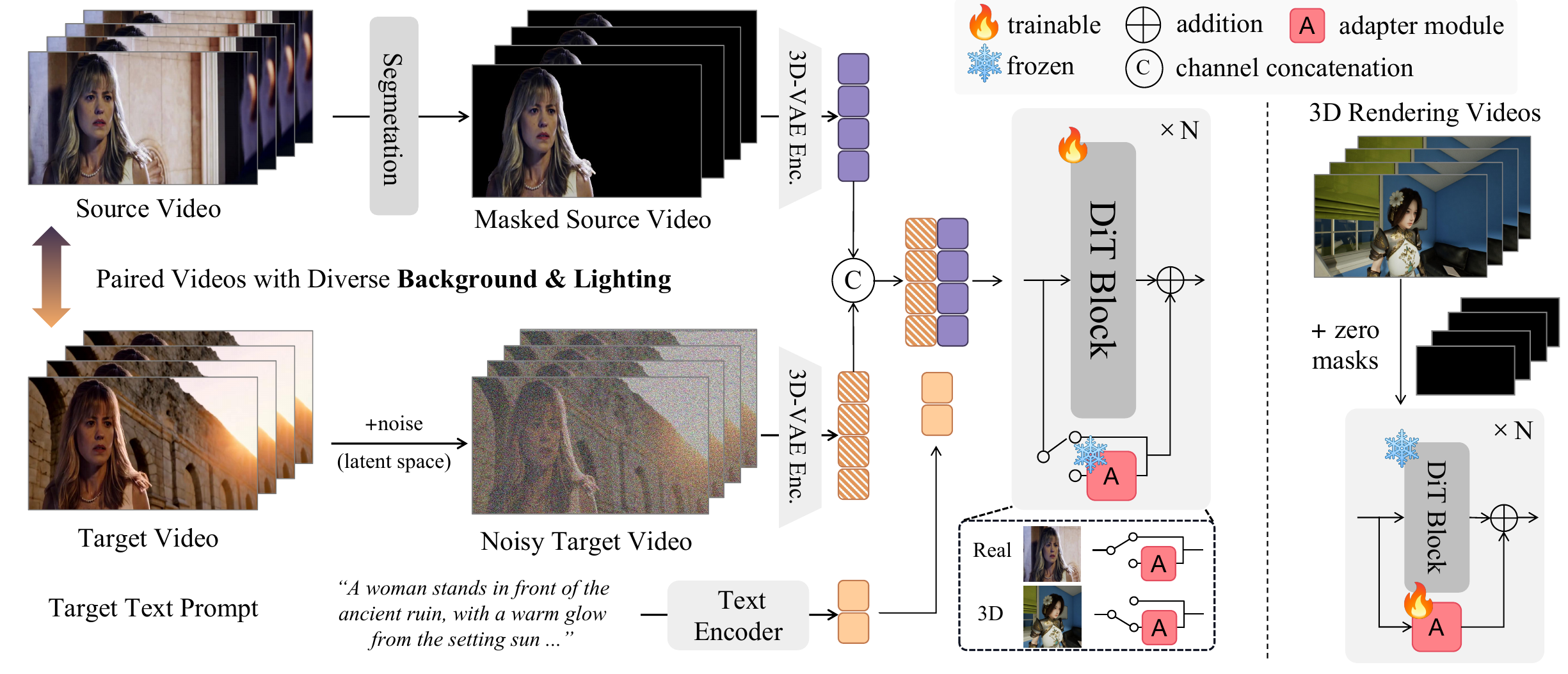}
\caption{\textbf{The framework of \method}, which is developed on a video generative model in DiT architecture. The model consumes the concatenation of noisy tokens and the masked input video. An adapter module is injected into the backbone to decouple the style distribution in 3D paired videos. 
}
\label{fig:framework}
\end{figure}

\textit{Paired Video Rendering:}
To formulate paired relighting videos, we adjust the background conditions for each combination of character, animation and camera movement.
The background is modified by either setting the character in various scenes, or modulating the lighting configurations with different light colors and intensities.
Therefore, we could obtain a group of videos with aligned foreground and diverse backgrounds in each group.
Finally, by combining $15$ environments with $100$ selected scenes, $20$ characters, $20$ animations and $10$ camera movements, we construct the synthetic dataset with $20,000$ videos in resolution of $1920 \times 1080$ and frame length of $90$ (corresponding to $6$ seconds).
Since the videos are in groups, we could sample at least $100,000$ training pairs via randomly pick two videos inside each group.

\paragraph{\textbf{In-the-wild Data}}
\label{sec:In-the-wild Data}

To mitigate the domain-gap of 3D rendered data, we further prepare realistic videos with simulated relighting effects in the following steps:

\textit{Data Collection:}
We start by gathering high-qualified videos from open sourced websites (\textit{e.g.,} Pexels~\cite{pexels}) or internal datasets, with shorter spatial side greater than $720$ and temporal length greater than $2$ seconds. 
The videos are then filtered with a single character clearly visible in the foreground.
The final collection contains roughly $100,000$ videos for training.

\textit{HDR-based Lighting Simulation:}
Inspired by the image degradation proposed in IC-Light~\cite{iclight}, 
we design a strategy of video lighting simulation to prepare relighted video (degraded) from the original in-the-wild video.
%
During training, the degraded video serves as input control, while the original video serves as ground truth output.
Considering that only the foreground of input video is consumed by the model, we ignore the visual content in background and concentrate on the lighting adjustment.
In brief, we firstly convert the original video into ``uniform-lit'' appearance inspired by LuminaBrush \cite{luminabrush2024}, then modify the lighting of converted video based on random HDR maps, which can be formulated as follows:

\textbf{1.} The uniform-lit video indicates those captured in the environments with uniformly assigned light sources in white color. We first select a set of videos under the sufficient lighting environments from our real-world video dataset as uniform-lit videos.
Given the collect uniform-lit videos as ground truth, the model inputs are prepared with the next step.

\textbf{2.} To simulate lighting adjustment in videos, we first use a video normal extractor~\cite{normalcrafter_bin2025} to obtain temporally consistent normal maps. Then we synthesize random environment maps, and conduct HDR rendering on videos based on the extracted normals. 
Specifically, we randomly sample a point set $\mathcal{P}$ as the light sources in the environment. For each point $p \in \mathcal{P}$, we assign a random light color, denoted as $I_p = (r,g,b)$. Then for any point $v$ in the environment map, we assume the light color in $v$ as the assembling of all light points with cosine light falloff:
\begin{equation}
    I_v = \min(\Sigma_{p \in \mathcal{P}} \max(I_{p} \cos\theta_{\langle v,~p\rangle}, 0), 255),
    \label{eq.coslight}
\end{equation}
where $\langle v,p\rangle$ denotes the radian distance between $v$ and $p$ in the spherical coordinate.
We then render the relighted video by applying the environment map on the original video, where the environment map is fixed across all frames.

\textbf{3.} Based on the above lighting simulation, we first apply this method on the uniform-lit videos, and use the paired videos to finetune a uniform-lit restorer.
Then for all the collected videos, we extract their uniform-lit appearance with the restorer, and conduct the HDR-based relighting to obtain the final degraded videos for training.

\subsection{\method: Video Relighting Model}

In this section, we present the model architecture of \method, together with a Multi-domain Joint training Curriculum on the aforementioned relighting video pairs.

\paragraph{\textbf{Model Architecture.}}
Considering the end-to-end generation manner, the fundamental capability of backbone model is critical to the quality of relighted videos.
In this paper, we adopt the most recent video generative model, \textit{i.e.,} Wan2.1~\cite{wan2025}, due to its remarkable performance with high fidelity and realistic details.
The model is based on diffusion transformer (DiT) architecture~\cite{dit}.
The overall architecture of \method is illustrated in \ref{fig:framework}.
Suppose the training video pair is $(V_{src},V_{tar})$, representing the source video and target video, respectively, and the foreground mask video is $M_{fg}$.
During training, we first encode the videos into latent space~\cite{stablediffusion} with the 3D-VAE encoder $\mathcal{E(\cdot)}$, and encode the corresponding description of target video into text tokens, noted as $T$.
Following the flow-matching process~\cite{liu2022flow} adopted in Wan2.1, we sample a timestep $t \in [0,1]$ and standard Gaussian noise $\epsilon$.
Then the final input of model is:
\begin{equation}
    X = \text{concat}([\mathcal{E}(V_{tar})*(1-t)+\epsilon * t];~\mathcal{E}(V_{src} \odot M_{fg})),
    \label{eq.concat_x}
\end{equation}
where $\text{concat}()$ denotes channel-wise concatenation, and $\odot$ denotes point-wide multiplication.
The input projection layer of DiT is correspondingly extended to consume the additional channels in the concatenated input.
Then the video tokens $X$ are projected and fed into transformer blocks together with the text tokens $T$, which guide the model to generate harmonious background and lighting for the video. 
Suppose the model is $\mathcal{F}_{\Theta}(\cdot)$ with trainable parameters $\Theta$, then the final optimization target is:
\begin{equation}
\min_{\Theta} \mathbb{E} \| \epsilon - \mathcal{F}_{\Theta} (X, T, t) \|_2^2
\label{eq.loss_func}
\end{equation} 

\paragraph{\textbf{Multi-domain Joint Training.}}
Given the paired videos from synthetic and realistic domains, a naive baseline solution is to directly optimize the model with mixed sampling of all data.
To better investigate the effects of the two domain data, we first conduct exploratory experiments with models separately trained on each domain.
Experiments suggest that: \textbf{1.} the model trained on the 3D domain achieves the light harmonization between foreground and background but may generated characters with ``plastic'' effects (closer to rendering data); 
\textbf{2.} model trained on the realistic data maintains the intrinsic consistency of foreground, which, however, may be lack of lighting effects and disharmonious with the background.

To better leverage the strengths of each domain, we design a two-stage training framework.
Based on the observation on separate-domain experiments, the synthetic data is mainly restricted by the domain-gap problem.
Ignoring the domain distribution, model benefit from rendered data with diverse lighting and shadow effects following physical rules.
To address this problem, inspired by the Still-Moving~\cite{chefer2024still}, we introduce a style adapter (implemented as LoRA~\cite{hu2022lora}) to decouple the style distribution of rendered videos. The model is trained in the following two stages:

For the first stage, we individually optimize the style adapter on the rendered data.
We set the source video as all-zero map, \textit{i.e.,} $V_{src} = 0$. 
The input text is prepended with a prefix, ``This is a video rendered with a 3D unreal engine'', to distinguish the video style with better convergence.
Through this stage, the adapter learns to generate videos in the 3D rendering style according to text descriptions.

For the second stage, we mix the 3D rendered data and in-the-wild data in a ratio of 1:1 sampling.
We fully finetune the DiT model while freeze the style adapter in this stage.
During training, the adapter is activated only in the forward process of 3D data, and is ignored for realistic data.
Such switchable process enables the model to learn the mapping between rendered pairs, while the adapter serves to absorb information related to domain distribution.
Then during inference, we remove the adapter from the finetuned DiT model, thus it achieves natural output with better foreground alignment.

\subsection{Benchmark and Metrics}

To fully evaluate the relighting performance of \method, especially from quantitative aspect, we construct a comprehensive video relighting benchmark with diverse range of lighting and background, and design metrics to evaluate the model from multiple perspectives.

\paragraph{\textbf{Benchmark Constitution.}}
The benchmark contains data in two domains, targeting different evaluation principles:

\textbf{1.} Paired 3D videos: We split $100$ pairs with diverse characters and scenes from the synthetic dataset for evaluation. Since there are strictly aligned pairs, we could evaluate with metrics of similarity measurement, \textit{e.g.,} PSNR~\cite{psnr}, SSIM~\cite{ssim} and LPIPS~\cite{lpips}.

\textbf{2.} Paired realistic videos: 
Though it is hard to obtain realistic pairs with ground-truth alignment, we construct pseudo-pairs with stricter alignment in the following steps:
\textbf{(1)} Following the same architecture of \method, we train an additional first-frame conditioned relighting model, which requires the first frame of target video as input. The frame is injected in extended input channel of the model.
\textbf{(2)} Given a set of realistic videos, we use an in-house image relighting model to transfer their first frames. 
The relighted frame then guide the model to generate complete relighted video, forming the pseudo pairs.
\textbf{(3)} We manually filter the pairs based on the criteria of foreground preservation, background quality and lighting harmonization. The final subset contains $100$ video pairs.

\textbf{3.} Unpaired realistic videos:
We further select $100$ high-quality videos with diverse lighting conditions and backgrounds for direct relighting test. Towards this subset, we do not measure similarity in pairs, but evaluate the general quality of output videos based on metrics from the widely-adopted V-Bench~\cite{huang2024vbench}, such as temporal consistency and video aesthetics. Videos in this subset can be further classified into three categories: 70 videos containing only one character in a close-up view, $15$ videos with one character in relatively far view, and $15$ videos containing multiple characters.

\paragraph{\textbf{Intrinsic Consistency Evaluation.}}
Besides the existing metrics in the above evaluation subsets, we further propose a new evaluation principle, termed \textit{intrinsic consistency}, capable of assessing the foreground preservation without ground-truth pairs.
Intrinsic consistency concentrate on the fixed attributes of subjects that are not influenced by the environments, such as albedo and texture.
The measurement is based on the uniform-lit restorer, denoted as $\mathcal{U}(\cdot)$, in the data preparation pipeline.
Given the source video $V_{src}$, foreground mask $M_{fg}$ and generated video $\hat{V}_{tar}$, we measure the consistency of their foregrounds by
\begin{equation}
    \text{sim}(\mathcal{U}(V_{src}) \odot M_{fg},~ \mathcal{U}(\hat{V}_{tar}) \odot M_{fg}),
    \label{eq.intrinsic consistency}
\end{equation}
where $\text{sim}(\cdot, \cdot)$ can be implemented as any similarity measuring metrics like PSNR, SSIM, and LPIPS.

\section{Experiments}

\subsection{Experimental Setup}

\paragraph{\textbf{Implementation Details.}}
We adopt a variant of Wan2.1, \textit{i.e.,} the controllable generative model Wan2.1-Fun-Control~\cite{VideoX_fun2025}, as our backbone. 
We train the model on 4 GPUs with batch size as $1$ and gradient accumulation steps as $2$, \textit{i.e.,} the actual batch size is $8$. 
The model is trained on the NVIDIA H200 GPUs.
The learning rate is set to $1^{-5}$ when fully fine-tuning the main DiT blocks, and $1^{-4 }$ when training the LoRA module.
The LoRA rank and $\alpha$ is set to be 16 and 16, and we inject LoRA into the input projection layer, QKV projection layer, output layer in attention and linear layers in the feed-forward network in all blocks.
The training steps are set to 40,000 for the first stage and 80,000 for the second. 
We use RMBG-2.0~\cite{rmbg2_0} to extract the foreground mask, and Qwen2.5-VL-7B-Instruct~\cite{Qwen2_5_VL} to caption the videos.

\paragraph{\textbf{Evaluation Metrics.}}
We use the PSNR~\cite{psnr}, SSIM~\cite{ssim}, LPIPS~\cite{lpips} to calculate the similarity between two videos, and use CLIP-T Score~\cite{clip_radford2021} to evaluate the alignment between the text prompt and the generated video. Several metrics in Vbench~\cite{huang2024vbench} (\textit{e.g.,} subject consistency, background consistency, motion smoothness, temporal flickering, etc.) are used to evaluate the overall quality of the generated video. The \textit{Intrinsic Consistency} metric (Eq.~\ref{eq.intrinsic consistency}) is used to evaluate the foreground preservation of the relighted videos when there is no available corresponding ground truth.

\paragraph{\textbf{Compared Methods.}}
We compare with existing methods supporting text-guided relighting and background replacement.
IC-Light~\cite{iclight} is a representative method for diffusion-based end-to-end image relighting. We adapt it to video relighting by applying it frame by frame. Light-A-Video~\cite{light_a_video_zhou2025} extend IC-Light to video relighting with training-free adaption, which designs a Consistent Light Attention (CLA) module and a Progressive Light Fusion (PLF) strategy to achieve stable video relighting results. 
The training-based RelightVid~\cite{RelightVid_fang2025} is not included since its text-guided relighting is not open-sourced.

\subsection{Performance on Video Relighting}

\begin{figure*}
  \centering
  \includegraphics[width=1.0\linewidth]{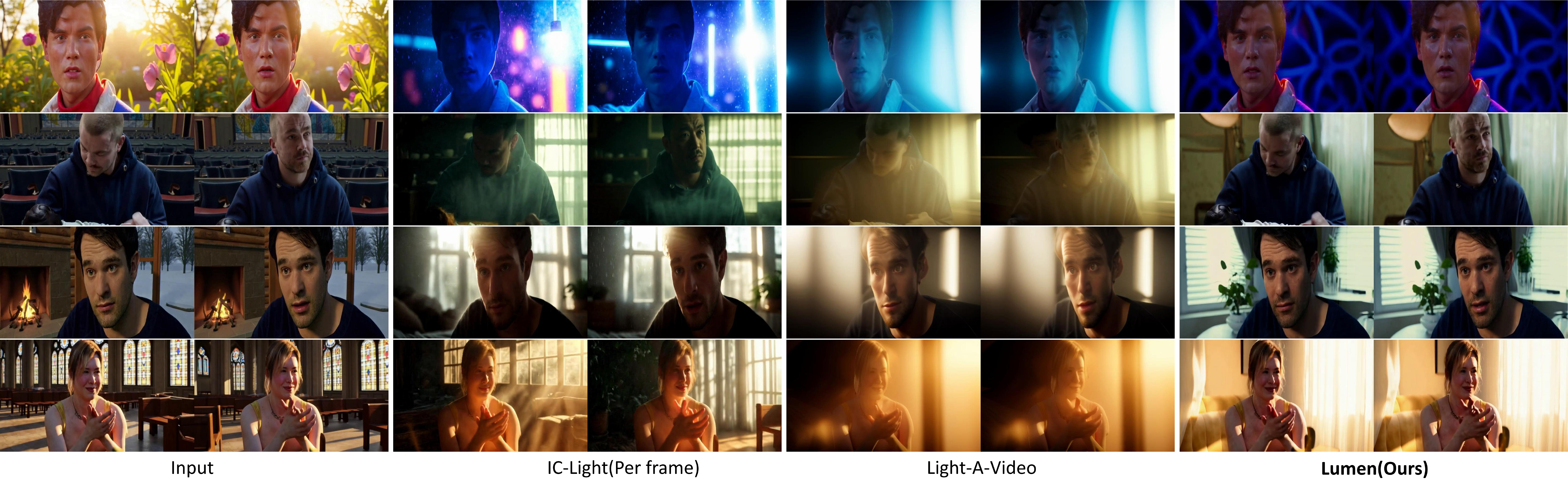}
  \caption{ \textbf{Qualitative Comparison}: We qualitatively compare \method with existing methods on the constructed benchmark. 
  The text prompts of the four cases from upper to lower are: (1) a man in a dimly lit room with blue lighting, neon lighting; (2) a man in the room, natural lighting, Wong Kar-wai style; (3) a man in a room, natural lighting; (4) a woman clapping hands, natural lighting, warm atmosphere.
  }
  \label{fig:compare}
\end{figure*}

\paragraph{\textbf{Qualitative Evaluation}}
Figure \ref{fig:compare} shows the qualitative results of \method and existing methods, which shows, Light-A-Video and IC-Light get poor results on the foreground preservation and background quality, while ours achieves better results on foreground preservation, background quality, and lighting harmonization.

\begin{table}
\caption{Comparison with existing methods on \textbf{paired synthetic/realistic videos}. Best and second are in \textbf{bold} and \underline{underline}.}
\vspace{+5pt}
\centering
\label{tab:method_comp1}

{\small
\begin{tabular}{l@{\extracolsep{2pt}} c@{\extracolsep{2pt}}c@{\extracolsep{2pt}}c@{\extracolsep{2pt}} c@{\extracolsep{2pt}} c@{\extracolsep{2pt}}c@{\extracolsep{2pt}}c@{\extracolsep{2pt}}c@{\extracolsep{2pt}}}
  \toprule
  \multirow{3}[1]{*}{\textbf{Method}} &
  \multirow{3}[1]{*}{\textbf{PSNR $\uparrow $}} & 
  \multirow{3}[1]{*}{\textbf{SSIM $\uparrow $}} & 
  \multirow{3}[1]{*}{\textbf{LPIPS $\downarrow $}} & 
  \multirow{3}[1]{*}{\textbf{CLIP-T $\uparrow $}} & 
  \multicolumn{4}{c}{\textbf{VBench}} \\
  \cmidrule(lr){6-9}

  & & & & & 
  \textbf{Subject} & 
  \textbf{Background} &
  \textbf{Motion} & 
  \textbf{Temporal} \\
  & & & & & 
  \textbf{Consistency $\uparrow $} & 
  \textbf{Consistency $\uparrow $} &
  \textbf{Smoothness $\uparrow $} & 
  \textbf{Flickering $\uparrow $} \\

  \midrule
  \multicolumn{9}{c}{On the 3D paired videos} \\
  \midrule

  IC-Light       & 21.03          & 0.8856          & 0.1033       & \underline{0.2838} & 0.8524          & 0.8964               & 0.9284          & 0.9164          \\
  Light-A-Video  & \underline{22.34} & \textbf{0.9008} & \underline{0.0951} & 0.2793   & \underline{0.9564} & \textbf{0.9439}    & \underline{0.9832} & \textbf{0.9886} \\
  \method (Ours) & \textbf{22.39} & \underline{0.8985} & \textbf{0.0741} & \textbf{0.3395} & \textbf{0.9575} & \underline{0.9279} & \textbf{0.9885} & \underline{0.9648} \\

  \midrule
  \midrule
  \multicolumn{9}{c}{On the realistic paired videos} \\
  \midrule

  IC-Light &       18.96          & 0.8060      & \underline{0.1712} & \underline{0.3080} & 0.9257       & 0.9428             & 0.9636            & 0.9579          \\
  Light-A-Video &  \underline{19.41} & \underline{0.8329} & 0.1717   & 0.2978          & \textbf{0.9811} & \underline{0.9636} & \underline{0.9933} & \underline{0.9901}  \\
  \method (Ours) & \textbf{23.06} & \textbf{0.8620} & \textbf{0.1083} & \textbf{0.3214} & \underline{0.9808} & \textbf{0.9638} & \textbf{0.9943} & \textbf{0.9905} \\

\bottomrule
\end{tabular}
}
\end{table}

\paragraph{\textbf{Quantitative Evaluation}}
The evaluation is conducted on paired synthetic videos, paired realistic videos and unpaired realistic videos.

\textit{Paired Videos Evaluation.}
We compare \method with existing methods on the paired 3D videos and paired realistic videos in our constructed benchmark.
The quantitative results are shown in Table \ref{tab:method_comp1}. 
When testing on paired videos, the similarity metrics (PSNR, SSIM, LPIPS) are calculated between the generated video and the ground truth video of the foreground region, indicating the foreground preservation of the generated video. 
The CLIP-T and VBench metrics are used to evaluate the text alignment and overall quality.

\textit{Unpaired Videos Evaluation.}
We compare \method with existing methods on unpaired realistic videos. The quantitative results are shown in Table \ref{tab:method_comp2}. 
When testing on unpaired videos, there is no ground truth available, so we use the proposed Intrinsic Consistency method to extract the uniform-lit of the foreground of the source video and the generated video, and then calculate the similarity between them, as formulated in Eq. \ref{eq.intrinsic consistency}. 

Results on both paired videos and unpaired videos show that \method outperforms existing methods in most metrics, indicating that our method can achieve better foreground preservation, background quality and lighting harmonization.

\begin{table}
\caption{Comparison with existing methods on \textbf{unpaired realistic videos}.}
\centering
\label{tab:method_comp2}
%
{\small
\begin{tabular}{l ccc c cccc}
  \toprule
  \multirow{3}[1]{*}{\textbf{Method}} &
  \multicolumn{3}{c}{\textbf{Intrinsic Consistency}} &
  \multirow{3}[1]{*}{\textbf{CLIP-T $\uparrow $}} &
  \multicolumn{4}{c}{\textbf{VBench}} \\
  \cmidrule(lr){2-4} \cmidrule(lr){6-9}
  
  & \multirow{2}[1]{*}{\textbf{PSNR $\uparrow $}} & 
  \multirow{2}[1]{*}{\textbf{SSIM $\uparrow $}} & 
  \multirow{2}[1]{*}{\textbf{LPIPS $\downarrow $}} & 
  &
  \textbf{Subj.} & 
  \textbf{Backg.} &
  \textbf{Mot.} & 
  \textbf{Tempo.} \\

  &  & & &  & 
  \textbf{Cons. $\uparrow $} & 
  \textbf{Cons. $\uparrow $} &
  \textbf{Smo. $\uparrow $} & 
  \textbf{Flic. $\uparrow $} \\

  \midrule
  IC-Light      & 18.42          & 0.8336          & 0.1210          & 0.1887          & 0.9667          & 0.9536          & 0.9751          & 0.9704          \\
  Light-A-Video & 20.16          & 0.8800          & 0.1141          & 0.1857          & 0.9882          & \textbf{0.9676} & 0.9941          & 0.9910          \\
  \method (Ours) & \textbf{23.55} & \textbf{0.9052} & \textbf{0.0650} & \textbf{0.2342} & \textbf{0.9909} & 0.9673          & \textbf{0.9953} & \textbf{0.9916} \\

  \bottomrule
\end{tabular}
}
\end{table}

\textit{User Study.}
We conduct a user study to evaluate the quality of the generated videos on the unpaired realistic videos from the following three aspects: 
1. \textbf{F}oreground \textbf{P}reservation: The consistency of the intrinsic attributes of the foreground subjects;
2. \textbf{B}ackground \textbf{Q}uality: The realism, richness and consistency of the background;
3. \textbf{L}ighting \textbf{H}armonization: The lighting consistency and harmonization between the foreground and background;
About 10 participants are invited to evaluate the results from the above three aspects, with score $\{1,2,3\}$ indicating bad, medium and good, respectively. The averaged and normalized scores are shown in Table \ref{tab:user_study}.
The results show that \method achieves the best scores on all three aspects by a large margin, indicating that our method can achieve remarkable performance on these three aspects.

\begin{table}[htbp]
\caption{The \textbf{user study} results on unpaired realistic videos.}
\centering
\label{tab:user_study}
\small
\begin{tabular}{lcccc}
  \toprule
  \multirow{3}[1]{*}{\textbf{Method}} &
  \multicolumn{4}{c}{{\textbf{User Study}}} \\
  \cmidrule(lr){2-5}
  &
  \textbf{Foreground } &
  \textbf{Background } &
  \textbf{Lighting } &
  \multirow{2}[1]{*}{\textbf{Average} $\uparrow$} \\
  & 
  \textbf{Preservation $\uparrow $} &
  \textbf{Quality $\uparrow $} &
  \textbf{Harmonization $\uparrow $} & \\

  \midrule
  IC-Light & 0.7567 & 0.7867 & 0.8300 & 0.7911 \\
  Light-A-Video & 0.8100 & 0.7433 & 0.8567 & 0.8033 \\
  \method (Ours) & \textbf{0.9133} & \textbf{0.9267} & \textbf{0.9533} & \textbf{0.9311} \\

  \bottomrule
\end{tabular}
\end{table}

\subsection{Ablation Studies}

We conduct ablation studies to evaluate the effectiveness of our Multi-domain Joint Training curriculum.
When training, we first train a LoRA branch on 3D rendered data, and then mix the 3D data and augmented in-the-wild data for joint training via a data-aware domain distillation method. 
Thus, we compare the results of the following four training methods:
1. \textbf{Only 3D Data}: Fully fine-tuning the main DiT blocks with only 3D rendered data;
2. \textbf{Only Real Data}: Fully fine-tuning the main DiT blocks with only augmented in-the-wild data;
3. \textbf{Mixed Data w/o Adapter}: Fully fine-tuning the main DiT blocks with 3D rendered data and augmented in-the-wild data;
4. \textbf{Mixed Data with Adapter}: Our Multi-domain Joint training curriculum with a domain-aware adapter.

\begin{figure}
  \centering
  \includegraphics[width=0.95\linewidth]{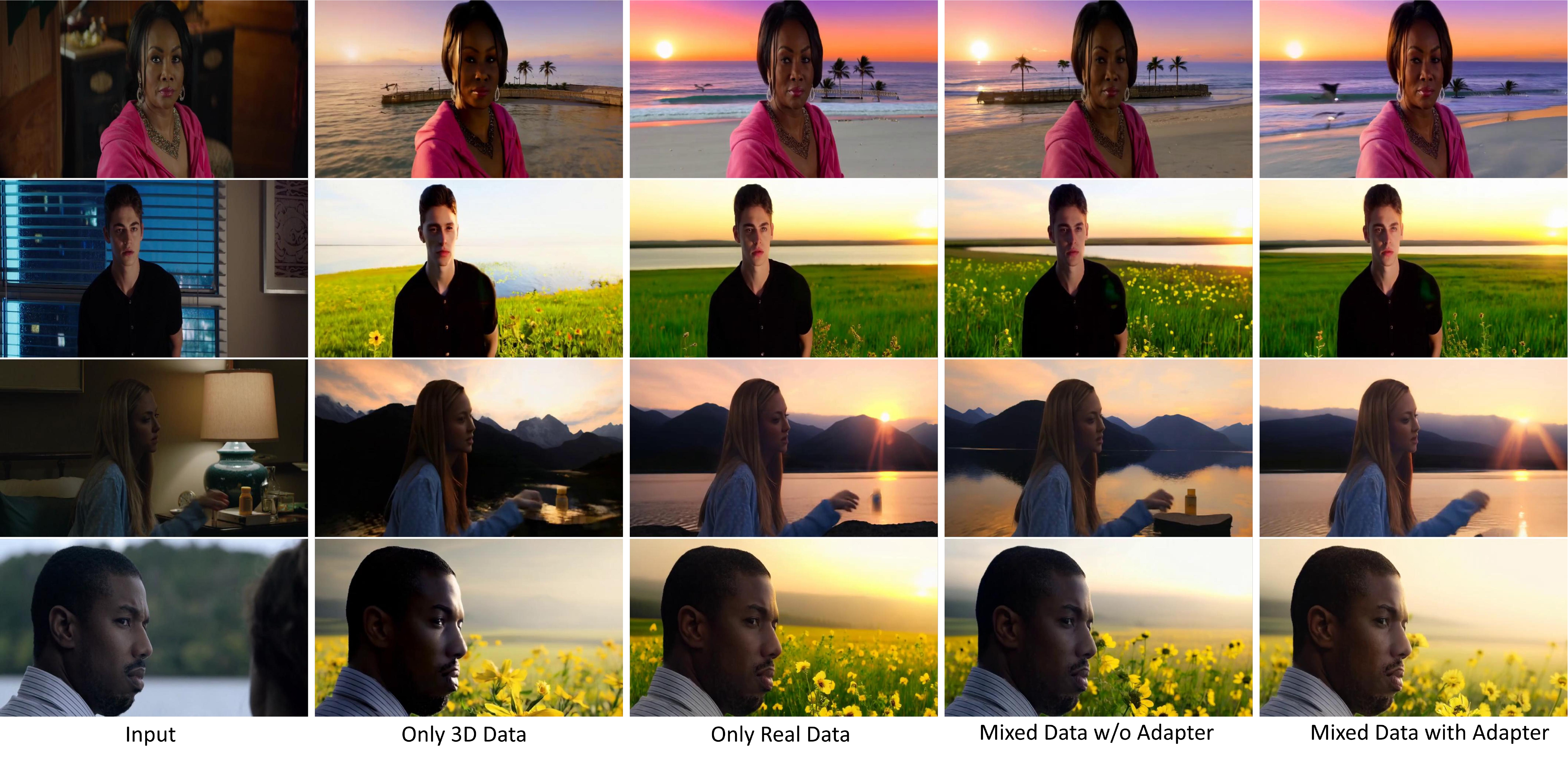}
  \caption{ \textbf{Qualitative Ablation Study}: Results of \textit{Only-3D-data} have better lighting effects but poor foreground preservation. Results of \textit{Only-real-data} may get unsatisfying lighting consistency. \method (Mixed Data with Adapter) get best results on both foreground preservation and lighting harmonization.
  }
  \label{fig:ablation}
\end{figure}

The qualitative ablation results are shown in Figure \ref{fig:ablation}. 
From the visiualization comparison, the results of \textit{Only-3D-data} have better lighting effects under the given background environment, but the foreground preservation is poor. 
Results of \textit{Only-real-data} may get unsatisfying lighting consistency between the foreground and background. 
Simply mixing the 3D data and real data for training can not achieve good results. 
Our multi-domain joint training curriculum with a domain-aware adapter achieves the best results on both foreground preservation and lighting harmonization.

The quantitative ablation results are shown in Table \ref{tab:ablation}. 
Using realistic data enhances the prompt following and video quality due to its generalized distribution compared with 3D data.
Directly mixing the two domains shows further improvement on prompt following, but also slightly decreases the quality.
Our final solution achieves balanced improvement on prompt following and video quality.
We note that quantitative metrics could not fully reflect the relighting performance, thus it is better to combine the results together with visualization for analysis.

\begin{table}
\caption{\textbf{Ablation study} of Multi-domain Joint training method on unpaired realistic videos. Best and second best results are in \textbf{bold} and \underline{underline}.}
\centering
\small
\label{tab:ablation}
\begin{tabular}{l c cccc}
  \toprule
  \multirow{2}[1]{*}{\textbf{Method}} &
  \multirow{2}[1]{*}{\textbf{CLIP-T $\uparrow $}} & 
  \multicolumn{4}{c}{\textbf{VBench}} \\
  \cmidrule(lr){3-6}
  & & \textbf{Subj. Cons. $\uparrow $} & 
  \textbf{Backg. Cons. $\uparrow $} &
  \textbf{Mot. Smo. $\uparrow $} & 
  \textbf{Tempo. Flic. $\uparrow $} \\

  \midrule
  Only 3D Data     & 0.2260           & 0.9879          & \textbf{0.9695} & 0.9936          & 0.9877          \\
  Only Real Data  & 0.2323           & \textbf{0.9916} & 0.9670          & 0.9946          & \underline{0.9907}          \\
  Mixed w/o Adapter  & \textbf{0.2377}  & \underline{0.9909} & 0.9667          & \underline{0.9950} & 0.9885          \\
  Mixed w/ Adapter & \underline{0.2342}  & \underline{0.9909} & \underline{0.9673} & \textbf{0.9953} & \textbf{0.9916} \\ 
  \bottomrule
\end{tabular}
\end{table}

\section{Conclusion}
In this paper, we present \method, a video relighting model based on large-scale video generative models, supporting text guidance on the lighting and background replacement.
To address the problem of lacking paired relighting videos, we construct a multi-domain dataset containing both 3D rendered and augmented realistic videos.
We adapt the pretrained video generative model by extending its input with the foreground of input video.
To fully leverage the constructed dataset, we design a domain mixed training curriculum, where a style adapter module is injected to help decoupling the domain distribution of rendered data.
Furthermore, we present a comprehensive benchmark with paired/unpaired videos in synthetic/realistic domain, together with new metrics examining the foreground preservation results.
Both qualitative and quantitative comparison demonstrate the effectiveness of \method compared with existing methods, where \method generates cinematic videos with consistent foreground preservation and temporal relighting.

\begin{figure*}[htbp]
  \centering
  \includegraphics[width=1.0\linewidth]{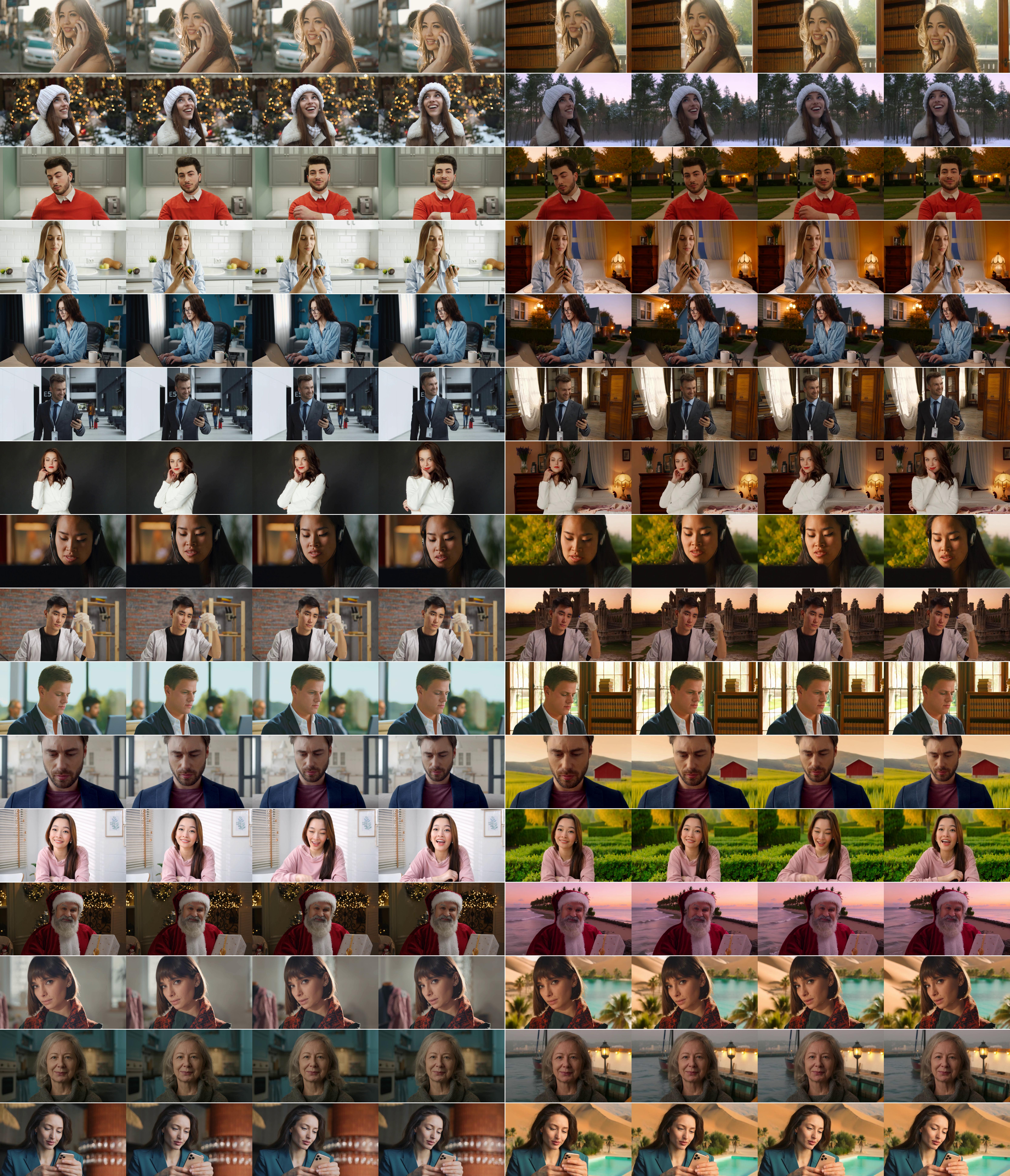}
  \caption{ \textbf{Video relighting results of \method} }
\end{figure*}

\begin{figure*}[htbp]
  \centering
  \includegraphics[width=1.0\linewidth]{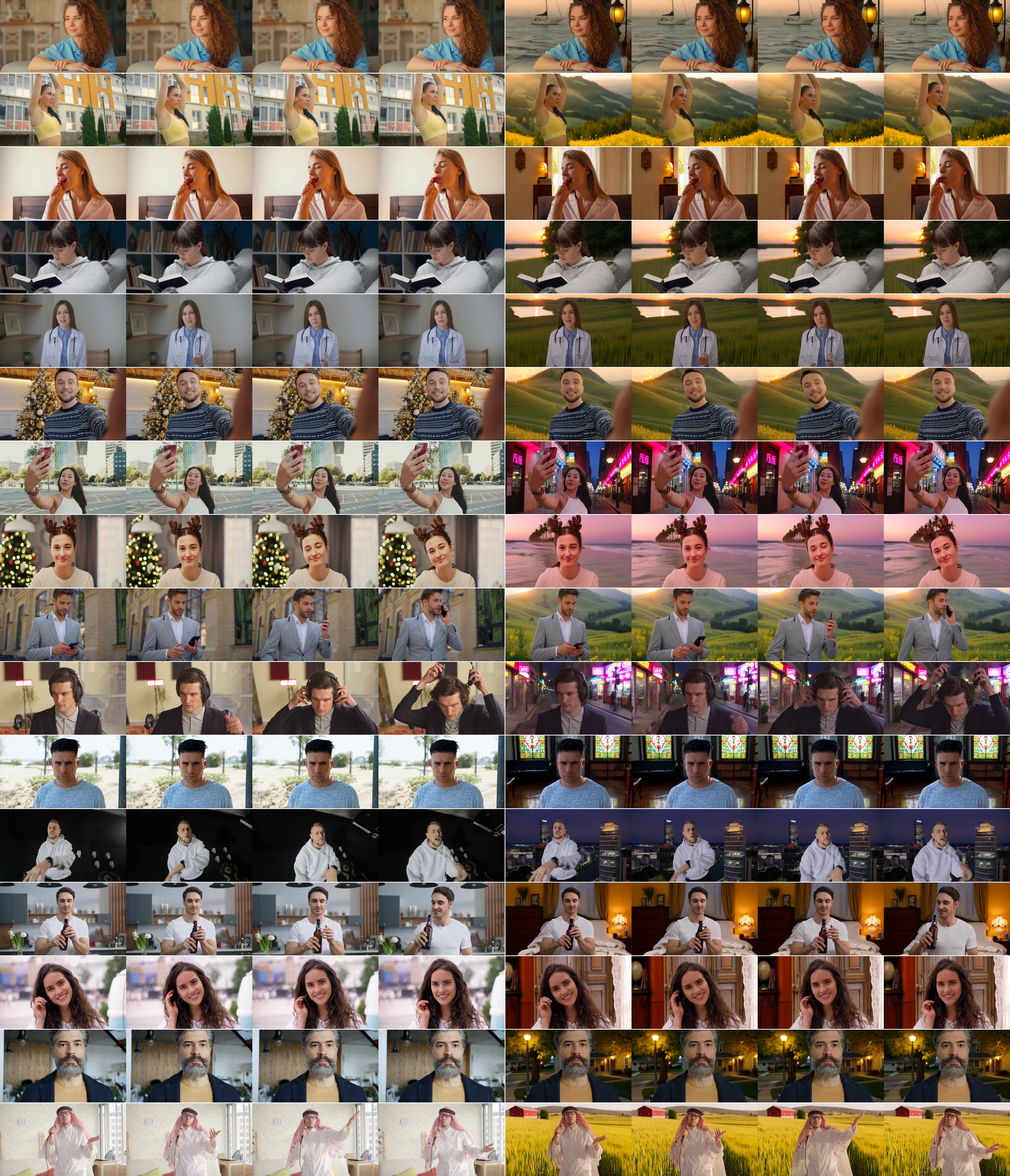}
  \caption{ \textbf{Video relighting results of \method} }
\end{figure*}

\clearpage 

\bibliographystyle{abbrvnat}
\bibliography{ref}

\end{document}